\documentclass[12pt]{article}

\usepackage[utf8]{inputenc}
\usepackage[T1]{fontenc}
\usepackage{amsmath,amssymb}
\usepackage{graphicx}
\usepackage{booktabs}
\usepackage{multirow}
\usepackage{hyperref}
\usepackage{natbib}
\usepackage[margin=1in]{geometry}
\usepackage{authblk}
\usepackage{caption}
\usepackage{array}
\usepackage{tabularx}

\hypersetup{
    colorlinks=true,
    linkcolor=blue,
    citecolor=blue,
    urlcolor=blue
}

\title{AI Psychometrics: Evaluating the Psychological Reasoning of Large Language Models with Psychometric Validities}

\author[1]{Yibai Li \thanks{yibai.li@scranton.edu}}
\author[2]{Xiaolin Lin \thanks{xiaolin.lin@csus.edu}}
\author[3]{Zhenghui Sha \thanks{zsha@austin.utexas.edu}}
\author[4]{Zhiye Jin \thanks{zjin@m.marywood.edu}}
\author[1]{Xiaobing Li \thanks{xiaobing.li@scranton.edu}}

\affil[1]{The University of Scranton}
\affil[2]{California State University, Sacramento}
\affil[3]{The University of Texas at Austin}
\affil[4]{Marywood University}

\date{}

\begin{document}

\maketitle

\begin{abstract}
The immense number of parameters and deep neural networks make large language models (LLMs) rival the complexity of human brains, which also makes them opaque ``black box'' systems that are challenging to evaluate and interpret. AI Psychometrics is an emerging field that aims to tackle these challenges by applying psychometric methodologies to evaluate and interpret the psychological traits and processes of artificial intelligence (AI) systems. This paper investigates the application of AI Psychometrics to evaluate the psychological reasoning and overall psychometric validity of four prominent LLMs: GPT-3.5, GPT-4, LLaMA-2, and LLaMA-3. Using the Technology Acceptance Model (TAM), we examined convergent, discriminant, predictive, and external validity across these models. Our findings reveal that the responses from all these models generally met all validity criteria. Moreover, higher-performing models like GPT-4 and LLaMA-3 consistently demonstrated superior psychometric validity compared to their predecessors, GPT-3.5 and LLaMA-2. These results help to establish the validity of applying AI Psychometrics to evaluate and interpret large language models.

\end{abstract}

\section{Introduction}

GPT-4, the latest iteration of OpenAI's language models, is reported to have around 1.76 trillion parameters \citep{Max2023}, making it a vast leap from GPT-3's 175 billion parameters. The complexity of these models stems from their vast number of parameters and deep neural networks, which facilitate a nuanced understanding and generation of human-like text. However, the sheer number of parameters and their intricate interconnections mean that even the developers of these models cannot predict or easily interpret why a model exhibits certain behaviors \citep{Ornes2023}. The complex computational architecture makes it nearly impossible to trace the path from input to output. This ``black box'' nature \citep{Woodside2024} presents challenges not only for improvements and troubleshooting but also raises ethical concerns regarding accountability and transparency in AI deployment.

Some researchers suggest that, given the complexity of today's large language models (LLMs), which rivals that of the human brain, we might benefit from applying human brain study techniques, such as psychometrics, to these AI systems. Psychometrics assesses psychological traits like intelligence, personality, and aptitude, providing a structured approach to evaluating LLMs' capabilities and behaviors. Building on this idea, these researchers have introduced the concept of AI Psychometrics. AI Psychometrics is an emerging field that applies psychometric methods to understand and measure the psychological traits and processes of artificial intelligence systems, as discussed by \citet{Wang2023} and \citet{Pellert2023}.

AI Psychometrics can be utilized to assess the psychological reasoning capabilities of large language models (LLMs). Current evaluations of LLMs predominantly focus on knowledge testing, logical reasoning, mathematical reasoning, and other task-based assessments \citep{Guo2023}. These evaluations often prioritize traditional intelligence tests (IQ) while largely neglecting assessments of emotional intelligence (EQ), such as psychological reasoning abilities. However, psychological reasoning---the ability to comprehend and anticipate thoughts, emotions, intentions, and human behaviors---is critically important for the development of artificial general intelligence (AGI) \citep{McLean2023}.

AI psychometrics is still in its nascent stage, primarily exploring preliminary ideas and concepts. Nonetheless, the application of traditional psychometric tests to large language models (LLMs) raises significant questions about their validity and reliability. To assess the validity of AI psychometrics, we hypothesized that LLMs would exhibit robust convergent, discriminant, predictive, and external validity in response to psychometric assessments.

We specifically tested the well-established Technology Acceptance Model (TAM) on four LLMs from two different families: GPT-3.5 and GPT-4o from OpenAI; and LLaMA-2, LLaMA-3 from Meta. Our findings indicate that these models generally met all validity criteria, except for LLaMA-2, which fell short in convergent validity. Moreover, higher-performing models like GPT-4o and LLaMA-3 consistently demonstrated superior psychometric validity compared to their predecessors, GPT-3.5 and LLaMA-2.

\section{AI Psychometrics: Definitions}

AI Psychometrics \citep{Wang2023,Pellert2023} is a field of study focused on applying psychometric methodologies to evaluate and interpret the psychological traits and processes of artificial intelligence (AI) systems. This discipline employs methods such as questionnaires, standardized tests, experiments, behavioral observations, interviews, and various statistical analyses to facilitate a deeper understanding of the cognitive processes of AI systems and their human-like properties. The goal of AI Psychometrics is to provide insights into complex AI systems, thereby enhancing their development, deployment, and management.

In the literature, a related concept is Psychometric AI \citep{Bringsjord2011}. This refers to AI systems that integrate psychometric models directly into their development, enabling them not only to perform well on a broad spectrum of intelligence and mental ability tests, but also allowing these systems to behave more like humans and achieve what humans can outside of these assessment settings \citep{Bringsjord2003}.

Another distinct but related field is Computational Psychometrics \citep{vonDavier2017}. It involves applying computational methods to traditional psychometrics, extending its capabilities through data mining, simulations, and AI algorithms. These methods analyze new types of data collected through sensors, eye-trackers, smartwatches, and other digital devices. For example, a study by \citet{WangEI2023} that uses eye-tracking technology to measure Emotional Intelligence (EI) falls into the category of Computational Psychometrics.

\section{Literature Review}

The integration of psychometrics and artificial intelligence (AI) can trace its conceptual origins back to the early years of AI in the 1950s. Thomas G. Evans was a pioneer in this field, introducing a computer program in 1964 capable of solving geometrical analogy tasks from intelligence tests developed in the 1940s. This early exploration underscored AI's potential to engage with established psychological assessments, setting a precedent for future research \citep{Evans1964,Bringsjord2011,Pellert2023}.

The discourse on AI's capacity to assimilate diverse experimental outcomes from psychology into a cohesive framework was notably advanced by \citet{Newell1973}. He proposed that AI should not only emulate human cognitive processes but also perform standard intelligence assessments such as the WAIS or Stanford-Binet. Newell's perspective highlighted the psychometric approach as a plausible method for this integration, suggesting a direct pathway for AI to mimic human intelligence.

By the early 2000s, the term ``psychometric AI'' was explicitly defined by \citet{Bringsjord2003} as a field devoted to constructing entities capable of performing well on a broad spectrum of intelligence and mental ability tests, beyond the narrow confines of traditional IQ assessments. This included the capacity for artistic and mechanical problem-solving, reflecting a more holistic view of intelligence.

The development of ``universal psychometrics'' by \citet{HernandezOrallo2012,HernandezOrallo2014} further expanded the ambit of this field, proposing methodologies to assess cognitive abilities across humans, animals, and machines. This approach aimed to create a unified framework for evaluating intelligence that integrates human psychometrics, comparative cognition, and AI metrics, thus bridging gaps between different cognitive entities.

The advent of large language models (LLMs) like ChatGPT in 2020 marked a significant evolution in AI research, presenting both opportunities and challenges for AI psychometrics. LLMs, with their vast parameter spaces, offer new vistas for understanding cognitive processes, yet they also pose challenges in evaluating such complex systems.

Current evaluations of LLMs primarily concentrate on knowledge testing and their abilities in logical reasoning, mathematical reasoning, and various tasks \citep{Guo2023}, akin to intelligence tests (IQ), but significantly overlook the assessment of emotional intelligence (EQ) such as psychological reasoning abilities. However, the ability to understand and predict thoughts, emotions, intentions, and human behaviors---key psychological reasoning capabilities---are more important for the development of artificial general intelligence (AGI). \citet{Wang2023} argue for a transition from task-oriented to construct-oriented evaluations, utilizing psychometric principles to probe the underlying psychological constructs of AI performance.

Regarding methodology, current evaluations of large language models (LLMs) rely primarily on basic psychometric applications, such as benchmark datasets and conventional intelligence tests. \citet{Wang2023} and \citet{Pellert2023} suggest using psychometrics to assess the psychological attributes and processes of LLMs. They argue that AI psychometrics provides a valuable framework for a deeper understanding of the cognitive mechanisms that underpin LLMs and also offers tools to analyze their psychological reasoning capabilities---an area that remains largely unexplored in existing research.

\citet{Pellert2023} have demonstrated the feasibility of applying psychometrics to assess the psychological profiles of LLMs using assessments like the Big Five Personality Traits, Dark Tetrad, Value Orientations, Moral Norms, and Gender/Sex Diversity Beliefs. However, both \citet{Wang2023} and \citet{Pellert2023} raise concerns about the validity and reliability of traditional psychometric tests when applied to LLMs. These models respond dynamically to changes in prompts; even minor changes can lead to inconsistent results, challenging the fundamental principles of psychometrics \citep{Wang2023}. They advocate for further validity research on the application of AI psychometrics in LLMs.

This study aims to address their call for research, focusing on the reliability and validity of psychometric evaluations in the context of large language models.

\section{Theories and Hypotheses}

In cognitive psychology, psychological reasoning refers to the cognitive processes involved in understanding and predicting the thoughts, emotions, intentions, and behaviors of oneself and others \citep{Scott1996}. It is a critical aspect of social cognition, enabling individuals to navigate social interactions and build relationships. Psychological reasoning is fundamental to human interaction and is studied extensively in fields such as psychology, cognitive science, and developmental psychology \citep{Baillargeon2014}.

In the context of large language models, the psychological reasoning capability of LLMs can be evaluated by how well they respond to psychological assessments of latent constructs and psychological models. This includes their ability to consistently and accurately mimic human responses to these assessments \citep{Wang2023,Shu2021}.

Validity theory is a field of study in psychometrics that provides useful theories, models, techniques, and assessment toolkits for testing validity. One important concept is latent constructs \citep{Furr2021}, which are theoretical constructs representing unobservable processes or attributes, such as intelligence, personality traits, or attitudes. Various types of validity are discussed in relation to latent constructs, including convergent validity, discriminant validity, predictive validity, and external validity.

Convergent validity \citep{Shrout2012} in psychometrics refers to the degree to which two measures that theoretically should be related are, in fact, related. Large language models (LLMs) are trained on a diverse corpus of human-generated text, encompassing discussions, opinions, reflections, and a spectrum of human emotions and psychological states. This training enables LLMs to generate responses that appear to understand these states, and if accurately calibrated, they could potentially reflect these processes. If LLM responses consistently align with established psychological measures, this could suggest that these responses exhibit convergent validity. Therefore, we hypothesize:

\medskip
\noindent\textbf{Hypothesis 1:} \textit{Responses generated by large language models demonstrate effective convergent validity for underlying latent constructs.}

\medskip
In psychometrics, discriminant validity \citep{Furr2021} measures how distinct a construct is from others, evidenced by weak correlations with unrelated constructs. Large Language Models (LLMs) are trained on diverse datasets containing various discussions, explanations, and contexts related to psychological constructs. This training enables LLMs to discern subtle differences between these constructs. When responding to inputs about psychological topics, LLMs generate text based on observed training patterns, thus demonstrating contextually relevant and distinct engagement with specific constructs. Therefore, we hypothesize:

\medskip
\noindent\textbf{Hypothesis 2:} \textit{Responses generated by large language models demonstrate effective discriminant validity for underlying latent constructs.}

\medskip
Predictive validity \citep{Rust2014} refers to a test's ability to forecast the outcomes it claims to measure. For language models, this involves predicting psychological traits by comparing the model's responses to human responses that correlate with specific psychological constructs. Large Language Models (LLMs) excel in pattern recognition, identifying and generating responses based on observed patterns during training. When presented with language indicative of psychological constructs such as emotions, personality traits, or cognitive biases, LLMs can reflect these constructs, having learned from similar patterns.

\medskip
\noindent\textbf{Hypothesis 3:} \textit{Responses generated by large language models demonstrate effective predictive validity for underlying latent constructs.}

\medskip
External validity \citep{Raykov2011} refers to the degree to which a study's findings can be generalized across various contexts and populations. A study---or in this context, a model's outputs---exhibits high external validity if its conclusions remain applicable in real-world scenarios outside the controlled conditions in which they were initially verified. For this study, external validity is achieved when the psychological models derived from responses produced by large language models (LLMs) align with those inferred from responses by human participants. This capability of LLMs stems from their training on diverse human-generated texts, allowing them to emulate aspects of human cognition, including heuristics, emotional expressions, and other psychological patterns. Therefore, we hypothesize:

\medskip
\noindent\textbf{Hypothesis 4:} \textit{Responses generated by large language models demonstrate effective external validity for underlying psychological models when compared to those generated by human participants.}

\medskip
High-performing large language models, typically characterized by their size and complexity, benefit from extensive and diverse training datasets. They undergo longer training periods, enabling them to generalize more effectively across a broad spectrum of topics and contexts. The psychological reasoning capability allows them to produce responses with a higher likelihood of maintaining validity across various psychometric assessments. These models are adept at interpreting and responding to subtle textual nuances that may elude less advanced models, thereby enhancing their psychometric validity. Therefore, we propose the following hypothesis:

\medskip
\noindent\textbf{Hypothesis 5:} \textit{Higher-performing large language models exhibit superior psychometric validity in their responses compared to those generated by lower-performing models.}

\medskip
In the next section, the methods of testing these hypotheses are discussed.

\section{Methodology}

\begin{figure}[htbp]
    \centering
    \includegraphics[width=0.7\textwidth]{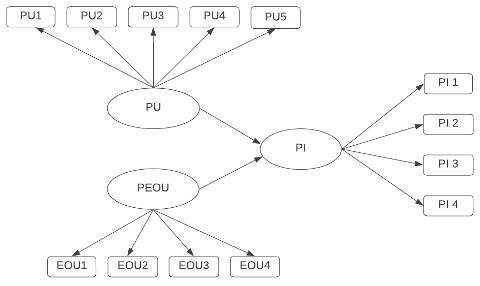} 
    \caption{Technology Acceptance Model}
    \label{fig:tam}
\end{figure}

To test these hypotheses, we employed the well-established Technology Acceptance Model (TAM) in the context of e-commerce. The specific structural equation model used in this study is depicted in Figure~\ref{fig:tam}.

TAM, originally developed by \citet{Davis1989}, predicts and explains user acceptance of information technology. It focuses on two primary factors: perceived usefulness and perceived ease of use. Perceived usefulness (PU) is defined as the degree to which a person believes that using a particular system would enhance their job performance or lead to other beneficial outcomes. Perceived ease of use (EOU) refers to the degree to which a person believes that using a particular system would be effortless. If a technology is easy to use, the user can operate it without difficulty, thereby reducing the cognitive strain involved in learning and using the technology.

The Technology Acceptance Model serves as a valuable framework for assessing the psychological reasoning capabilities of Large Language Models (LLMs) and the psychometric validities of their responses. (1)~The ability to understand the latent constructs of perceived usefulness and perceived ease of use, the relationship between the manifest variables and their respective latent constructs, and the ability to predict behavioral intentions are great indicators for LLMs' psychological reasoning capabilities. (2)~Another reason TAM serves the objectives of this study well is its widespread validation across different technologies and environments. The validity of the constructs and the measurement items, such as the ones we used in Table~\ref{tab:measurement}, has been extensively tested. (3)~The abundance of empirical results from human subjects makes it a robust model for comparing human responses to AI responses.

The survey is set in the context of evaluating consumer reactions to product recommendations on an online retail platform, Amazon. The focus of the survey is on understanding how users perceive these product recommendations as useful and easy to use and their intentions to continue purchasing products from the platform. The research instruments were carefully chosen and adapted from existing literature to ensure relevance and accuracy in measuring the constructs of interest. Specifically, items assessing perceived usefulness and ease of use were adapted from \citet{Davis1989}, tailored to fit the context of our study. Additionally, items measuring purchase intention were derived from \citet{Coyle2001}. All survey items were rated using a 7-point Likert scale, ranging from 1 (strongly disagree) to 7 (strongly agree), facilitating a standardized assessment of participant responses. The specific items and their sources are listed in Table~\ref{tab:measurement}.

\begin{table}[htbp]
\centering
\caption{Measurement Items}
\label{tab:measurement}
\small
\begin{tabular}{@{}ll@{}}
\toprule
\textbf{Item} & \textbf{Description} \\
\midrule
\multicolumn{2}{l}{\textit{Perceived Usefulness (PU)}} \\
PU1 & Product recommendations make my shopping experience more interesting. \\
PU2 & Product recommendations make my shopping experience more enjoyable. \\
PU3 & Product recommendations help me make more informed purchase decisions. \\
PU4 & Product recommendations help me make more accurate purchase decisions. \\
PU5 & Product recommendations meet my needs adequately in my purchase decisions. \\
\midrule
\multicolumn{2}{l}{\textit{Ease of Use (EOU)}} \\
EOU1 & It is easy to understand product recommendations. \\
EOU2 & It is easy to understand how to use product recommendations. \\
EOU3 & It is easy to learn how to use product recommendations. \\
EOU4 & It is easy to learn how product recommendations operate. \\
\midrule
\multicolumn{2}{l}{\textit{Purchase Intentions (PI)}} \\
PI1 & It is likely that I will continue purchasing products from Amazon. \\
PI2 & I will purchase products from Amazon next time I need them. \\
PI3 & Suppose that a friend wants my advice on his/her search for some products; \\
     & I would recommend him/her to purchase them from Amazon. \\
PI4 & I will definitely purchase products again from Amazon. \\
\bottomrule
\end{tabular}
\end{table}

\subsection{Diffusion Method}

One of the primary challenges in collecting responses for AI psychometrics in large language models (LLMs) is eliciting diverse answers. Typically, LLMs respond to prompts using the neural pathways with the highest weights, resulting in little variability. Our experiments show that even modifying a model's temperature or random seeds does not significantly increase variability in LLM responses, limiting their potential for statistical testing. To address this issue, we have adopted the diffusion method, a technique from the broader family of diffusion models inspired by physical processes such as the dispersion of gases \citep{SohlDickstein2015}. These models are renowned in machine learning for generating high-quality, diverse images \citep{Ho2020}. Typically, these models introduce noise to the system and operate recursively and sequentially, building on previous states to generate subsequent states of the system. In this study, we specifically apply the diffusion method by starting with a randomly selected question paired with a random answer, thus establishing the initial system state randomly. This method then collects extensive response sets from LLMs through iterative prompts, each building on the previous one. By capturing a broad spectrum of AI behaviors and responses, the diffusion method not only deepens the data pool but also enables further statistical analysis and psychometric evaluations.

\paragraph{Definitions.}
\begin{itemize}
    \item $Q$: Set of all survey questions, $Q = \{q_1, q_2, \ldots, q_n\}$.
    \item $A$: Set of 7-point Likert-scale answers for each question, $A = \{1, 2, 3, 4, 5, 6, 7\}$.
    \item History ($H$): A sequence of answered questions and their corresponding answers at any point, $H = \{(q_i, a_i), (q_j, a_j), \ldots\}$ where $q_i, q_j \in Q$ and $a_i, a_j \in A$.
\end{itemize}

\paragraph{Initial Step.}
\begin{enumerate}
    \item Select the first question $q_k$ randomly from $Q$.
    \item Randomly assign an answer $a_k$ from $A$.
    \item Update $H$ to include the initial question--answer pair: $H \leftarrow \{(q_k, a_k)\}$.
\end{enumerate}

\paragraph{Recursive Selection and Prediction.}
\begin{enumerate}
    \item Select the next question $q_n$ randomly from unanswered questions $Q' = Q \setminus \{q \mid (q, a) \in H\}$.
    \item Predict the answer $a_n$ for $q_n$ based on the history $H$:
    \begin{equation}
        a_n = \arg\max_{a \in A} P(a \mid q_n, H)
    \end{equation}
    \item Update $H$ to include the new question--answer pair: $H \leftarrow H \cup \{(q_n, a_n)\}$.
\end{enumerate}

\paragraph{Termination.} Continue the previous step until all questions in $Q$ are answered.

\subsection{Data Collection}

To assess the validity of responses from large language models, we selected two representative models from each of two distinct families. From OpenAI, we utilized GPT-3.5-turbo and GPT-4o, and from Meta, we used LLaMA-2-13B-chat and LLaMA-3-8B-instruct. These models were accessed through the OpenRouter API (\url{https://openrouter.ai/}), which facilitates streamlined integration and scalability for a wide variety of large language models. Employing the diffusion method described in the previous section, we queried each model 500 times under identical experimental conditions to ensure data consistency.

We extracted a single numerical ``answer value'' from each model per experiment, ensuring uniformity in data collection. These values were then ordered and analyzed to discern patterns and validate the model responses. This process generated four distinct datasets from the AI models, enabling a robust analysis of their response validity.

Parallel to AI data, we also collected responses from human participants to provide a comparative baseline. In May 2024, we conducted a structured survey through Amazon Mechanical Turk. Of the 286 initial participants, 248 successfully completed the survey, meeting all screening and quality assurance criteria. Eligibility for participation was contingent upon having made at least two purchases on Amazon within the preceding three months. Quality assurance measures included specific questions designed to verify participant attentiveness and to filter out careless responses. Each participant who completed the survey received a stipend as compensation.

\section{Data Analysis and Validity Assessment}

In our study, we applied Partial Least Squares-based Structural Equation Modeling (PLS-SEM) to evaluate items and test hypotheses. We utilized SmartPLS with a bootstrap resampling method involving 5,000 samples to assess the significance of the path coefficients. The data analysis was conducted independently for each of the five datasets, and the results are presented separately for each set.

\subsection{Convergent Validity}

Factor loadings are crucial in evaluating convergent validity. Convergent validity is confirmed when the loadings for a factor exceed a critical threshold, typically set at 0.50 as suggested by \citet{Hulland1999}, indicating that individual items correlate well with their respective constructs. The data collected from GPT-3.5, GPT-4o, LLaMA-3, and human subjects, as shown in Table~\ref{tab:loadings}, all demonstrate factor loadings above this threshold, affirming good convergent validity. However, LLaMA-2 presents an exception, with the loading of PI4 recorded at only 0.48, which does not meet the established criterion.

\begin{table}[htbp]
\centering
\caption{Factor Loadings}
\label{tab:loadings}
\small
\begin{tabular}{@{}lccccc@{}}
\toprule
 & GPT-3.5 & GPT-4o & LLaMA-2 & LLaMA-3 & Human \\
\midrule
PU1  & 0.90 & 0.93 & 0.70 & 0.86 & 0.90 \\
PU2  & 0.89 & 0.92 & 0.72 & 0.89 & 0.88 \\
PU3  & 0.91 & 0.92 & 0.72 & 0.84 & 0.89 \\
PU4  & 0.89 & 0.92 & 0.73 & 0.86 & 0.93 \\
PU5  & 0.92 & 0.94 & 0.75 & 0.82 & 0.90 \\
EOU1 & 0.85 & 0.90 & 0.76 & 0.83 & 0.84 \\
EOU2 & 0.89 & 0.92 & 0.64 & 0.83 & 0.83 \\
EOU3 & 0.90 & 0.91 & 0.71 & 0.84 & 0.83 \\
EOU4 & 0.87 & 0.86 & 0.72 & 0.85 & 0.85 \\
PI1  & 0.88 & 0.91 & 0.68 & 0.82 & 0.84 \\
PI2  & 0.83 & 0.88 & 0.66 & 0.81 & 0.85 \\
PI3  & 0.80 & 0.87 & 0.56 & 0.75 & 0.78 \\
PI4  & 0.82 & 0.87 & 0.48 & 0.77 & 0.80 \\
\bottomrule
\end{tabular}
\end{table}

Cronbach's alpha is a measure used in statistics to evaluate the internal consistency or reliability of a scale, often used in psychological tests and surveys. It provides an index ranging from 0 to 1, indicating how closely related a set of items are as a group. Generally, a Cronbach's alpha value higher than 0.70 is considered acceptable, indicating that the items measure the same underlying concept. In Table~\ref{tab:alpha}, GPT-3.5, GPT-4o, and LLaMA-3 have Cronbach's alpha scores that exceed this critical threshold. However, LLaMA-2 recorded lower Cronbach's alpha values such as 0.68 and 0.41, suggesting a weaker internal consistency in its responses to questions about Ease of Use (EOU) and Purchase Intentions (PI).

\begin{table}[htbp]
\centering
\caption{Cronbach's Alpha}
\label{tab:alpha}
\small
\begin{tabular}{@{}lccccc@{}}
\toprule
 & GPT-3.5 & GPT-4o & LLaMA-2 & LLaMA-3 & Human \\
\midrule
PU  & 0.94 & 0.96 & 0.77 & 0.91 & 0.83 \\
EOU & 0.90 & 0.92 & 0.68 & 0.86 & 0.86 \\
PI  & 0.85 & 0.91 & 0.41 & 0.79 & 0.83 \\
\bottomrule
\end{tabular}
\end{table}

Composite reliability (CR) is another measure widely used to assess the internal consistency of a latent variable model, reflecting how well the items of a scale work together to define an unobservable construct. The widely accepted cut-off value for composite reliability is 0.70. Table~\ref{tab:cr} shows that GPT-3.5, GPT-4o, LLaMA-2, LLaMA-3, and human all have scores above this threshold, indicating good reliability and convergent validity.

\begin{table}[htbp]
\centering
\caption{Composite Reliability}
\label{tab:cr}
\small
\begin{tabular}{@{}lccccc@{}}
\toprule
 & GPT-3.5 & GPT-4o & LLaMA-2 & LLaMA-3 & Human \\
\midrule
PU  & 0.96 & 0.97 & 0.84 & 0.93 & 0.95 \\
EOU & 0.93 & 0.94 & 0.80 & 0.90 & 0.90 \\
PI  & 0.90 & 0.93 & 0.90 & 0.87 & 0.89 \\
\bottomrule
\end{tabular}
\end{table}

Average Variance Extracted (AVE) measures the amount of variance that a latent variable captures from its indicators relative to the amount of variance due to measurement error. The generally accepted cut-off value for AVE is 0.50 \citep{Gefen2005}. This means that 50\% or more of the variance of the observed variables should be accounted for by the latent variable. Table~\ref{tab:ave} shows that GPT-3.5, GPT-4o, LLaMA-3, and human have AVE scores greater than 0.50. Only LLaMA-2's AVE score on purchase intention (PI) is below the threshold (0.36).

\begin{table}[htbp]
\centering
\caption{Average Variance Extracted}
\label{tab:ave}
\small
\begin{tabular}{@{}lccccc@{}}
\toprule
 & GPT-3.5 & GPT-4o & LLaMA-2 & LLaMA-3 & Human \\
\midrule
PU  & 0.81 & 0.86 & 0.52 & 0.73 & 0.81 \\
EOU & 0.77 & 0.80 & 0.51 & 0.70 & 0.70 \\
PI  & 0.70 & 0.78 & 0.36 & 0.62 & 0.67 \\
\bottomrule
\end{tabular}
\end{table}

In summary, Hypothesis~1 posits that responses generated by large language models effectively demonstrate convergent validity for underlying latent constructs. This hypothesis is supported for GPT-3.5, GPT-4, and LLaMA-3. However, it is not supported for LLaMA-2.

\subsection{Discriminant Validity}

One common method to assess discriminant validity is the Fornell--Larcker criterion \citep{Fornell1981}. This criterion states that discriminant validity is established if the square root of the Average Variance Extracted (AVE) of each construct is greater than the correlations between that construct and any other constructs in the model, which suggests that the construct is more closely related to its own indicators than to others, supporting discriminant validity. Tables~\ref{tab:disc_gpt}, \ref{tab:disc_llama}, and \ref{tab:disc_human} show that GPT-3.5, GPT-4o, LLaMA-2, LLaMA-3, and human participants all meet this criterion. Therefore, Hypothesis~2---which posits that responses generated by large language models effectively demonstrate discriminant validity for underlying latent constructs---is supported.

\begin{table}[htbp]
\centering
\caption{Construct Correlations and Square Roots of AVEs from GPT-3.5 and GPT-4o}
\label{tab:disc_gpt}
\small
\begin{tabular}{@{}lccc@{}}
\toprule
 & PU & EOU & PI \\
\midrule
PU  & \textbf{0.88}/\textbf{0.90} & & \\
EOU & 0.23/0.54 & \textbf{0.83}/\textbf{0.88} & \\
PI  & 0.33/0.53 & 0.42/0.62 & \textbf{0.90}/\textbf{0.93} \\
\bottomrule
\end{tabular}

\medskip
\footnotesize\textit{Note:} Values are displayed in the format GPT-3.5/GPT-4o. Bold numbers along the diagonal represent the square roots of the AVEs.
\end{table}

\begin{table}[htbp]
\centering
\caption{Construct Correlations and Square Roots of AVEs from LLaMA-2 and LLaMA-3}
\label{tab:disc_llama}
\small
\begin{tabular}{@{}lccc@{}}
\toprule
 & PU & EOU & PI \\
\midrule
PU  & \textbf{0.70}/\textbf{0.84} & & \\
EOU & 0.36/0.51 & \textbf{0.60}/\textbf{0.79} & \\
PI  & 0.50/0.67 & 0.40/0.59 & \textbf{0.72}/\textbf{0.85} \\
\bottomrule
\end{tabular}

\medskip
\footnotesize\textit{Note:} Values are displayed in the format LLaMA-2/LLaMA-3. Bold numbers along the diagonal represent the square roots of the AVEs.
\end{table}

\begin{table}[htbp]
\centering
\caption{Construct Correlations and Square Roots of AVEs from Human Participants}
\label{tab:disc_human}
\small
\begin{tabular}{@{}lccc@{}}
\toprule
 & PU & EOU & PI \\
\midrule
PU  & \textbf{0.84} & & \\
EOU & 0.75 & \textbf{0.82} & \\
PI  & 0.48 & 0.53 & \textbf{0.90} \\
\bottomrule
\end{tabular}

\medskip
\footnotesize\textit{Note:} Bold numbers along the diagonal represent the square roots of the AVEs.
\end{table}

\subsection{Predictive Validity}

Predictive validity refers to how well the model predicts a construct based on the interrelations among other constructs. $R^2$ values for endogenous constructs, like Purchase Intention (PI) in this study, serve as a key metric for assessing this aspect. A higher $R^2$ value indicates that the model explains a greater proportion of the variance in the predicted constructs, suggesting better predictive validity.

In this study, we examined the predictive validity of different LLMs (GPT-3.5, GPT-4o, LLaMA-2, LLaMA-3), comparing their performance against human subjects in predicting Purchase Intention (PI).

As indicated in Table~\ref{tab:path}, the $R^2$ value for human participants was the highest at 59.90\%. GPT-4 demonstrated the next highest level of predictive validity among the LLMs, with an $R^2$ value of 44.30\%, which is a substantial improvement over its predecessor, GPT-3.5, which had an $R^2$ value of only 18.40\%. LLaMA-3 also showed a relatively high $R^2$ of 37.30\%, outperforming LLaMA-2, which had an $R^2$ of 19.70\%. Overall, Hypothesis~3 is supported. Responses generated by large language models demonstrate considerable predictive validity for endogenous constructs, like Purchase Intention (PI).

\begin{table}[htbp]
\centering
\caption{Path Coefficients and the $R^2$ of Purchase Intention}
\label{tab:path}
\small
\begin{tabular}{@{}lccccc@{}}
\toprule
 & GPT-3.5 & GPT-4o & LLaMA-2 & LLaMA-3 & Human \\
\midrule
PU $\rightarrow$ PI  & $0.39^{***}$ & $0.46^{***}$ & $0.30^{***}$ & $0.46^{***}$ & $0.22^{***}$ \\
EOU $\rightarrow$ PI & $0.11^{*}$   & $0.30^{***}$ & $0.21^{***}$ & $0.19^{***}$ & $0.65^{***}$ \\
\midrule
$R^2$ of PI & 18.4\% & 44.3\% & 19.7\% & 37.3\% & 59.9\% \\
\bottomrule
\end{tabular}

\medskip
\footnotesize\textit{Note:} $^{*}p < 0.05$; $^{**}p < 0.01$; $^{***}p < 0.001$
\end{table}

\subsection{External Validity}

External validity refers to the extent to which the results of a study can be generalized beyond the specific conditions of the study. This includes generalizability to other populations, settings, times, and measures. In this study, external validity is established when the psychological process measured in the LLMs, in terms of path coefficients, align with established psychometric models responded by human subjects. Table~\ref{tab:path} shows the path coefficients from Perceived Usefulness (PU) to Purchase Intention (PI), and from Ease of Use (EOU) to Purchase Intention (PI), tested using GPT-3.5, GPT-4o, LLaMA-2, and LLaMA-3, are consistent with those tested using human subjects. They are all positively related with Purchase Intention, and they are all statistically significant. Therefore, Hypothesis~4---responses generated by large language models demonstrate effective external validity for underlying psychological models when compared to those generated by human participants---is supported.

\subsection{Summary of Results}

In summary, most of the proposed hypotheses regarding the psychometric validity of responses from large language models are supported. Responses generated by GPT-3.5, GPT-4o, LLaMA-2, and LLaMA-3 have demonstrated effective discriminant validity (Hypothesis~2), predictive validity (Hypothesis~3), and external validity (Hypothesis~4). However, only GPT-3.5, GPT-4o, and LLaMA-3, excluding LLaMA-2, have shown effective convergent validity (Hypothesis~1).

Additionally, higher-performing large language models such as GPT-4 and LLaMA-3 have consistently shown superior psychometric validity compared to those generated by GPT-3.5 and LLaMA-2. Table~\ref{tab:loadings} illustrates that the loadings from GPT-4 and LLaMA-3 are consistently higher than those from their respective predecessor models, GPT-3.5 and LLaMA-2. Tables~\ref{tab:alpha}, \ref{tab:cr}, \ref{tab:ave}, \ref{tab:disc_gpt}, and \ref{tab:disc_llama} demonstrate that GPT-4 and LLaMA-3 consistently achieve higher composite reliability, average variance extracted, Cronbach's alpha, and square root of AVE than their predecessors. In Table~\ref{tab:path}, the predictive power of GPT-4 and LLaMA-3, as indicated by $R^2$, is higher than that of GPT-3.5 and LLaMA-2, respectively. Therefore, Hypothesis~5 is supported in both the GPT and LLaMA model families in this study.

\section{Conclusions}

This study has demonstrated the feasibility and effectiveness of AI Psychometrics in assessing the psychological reasoning capabilities of large language models (LLMs). By applying the Technology Acceptance Model (TAM), we explored various dimensions of psychometric validity---convergent, discriminant, predictive, and external---across four prominent LLMs: GPT-3.5, GPT-4, LLaMA-2, and LLaMA-3. This study highlights that the application of rigorous psychometric techniques offers a promising avenue for deepening our understanding of AI's cognitive capabilities. This approach not only enhances the interpretability and transparency of AI systems but also helps in the responsible development \citep{Li2024} of AI systems that are capable of ethical and socially aware decisions, aligning more closely with human values and norms.

\bibliographystyle{plainnat}
\bibliography{references}

\end{document}